\pdfoutput=1

\documentclass[11pt]{article}

\usepackage[final]{acl}
\usepackage{times}
\usepackage{latexsym}
\usepackage{graphicx}
\usepackage{amsmath}
\usepackage{amssymb}
\usepackage{mathtools}
\usepackage{booktabs} 
\usepackage{soul}
\usepackage{tikz}
\usetikzlibrary{patterns}
\usepackage{pgfplots}
\pgfplotsset{compat=1.18}
\tikzset{font=\scriptsize}
\usepackage{colortbl}
\usepackage{array}
\usepackage{xcolor}
\usepackage{xspace}
\usepackage{multirow}
\usepackage{subfig}

\definecolor{leftgreen}{RGB}{158, 210, 190} 
\definecolor{midgreen}{RGB}{126, 170, 146}
\definecolor{rightred}{RGB}{255, 217, 183}

\definecolor{napiergreen}{rgb}{0.16, 0.5, 0.0}
\definecolor{parisgreen}{rgb}{0.31, 0.78, 0.47}

\definecolor{fluorescentorange}{rgb}{1.0, 0.75, 0.0}
\definecolor{orange(colorwheel)}{rgb}{1.0, 0.5, 0.0}

\definecolor{Blueback}{RGB}{218, 227, 243} 
\definecolor{Greenback}{RGB}{226, 240, 217}
\definecolor{Redback}{RGB}{251, 229, 214}

\usepackage[T1]{fontenc}

\usepackage[utf8]{inputenc}

\usepackage{microtype}

\usepackage{inconsolata}

\newcommand*\samethanks[1][\value{footnote}]{\footnotemark[#1]}
\title{Prompt Chaining or Stepwise Prompt? Refinement in Text Summarization}

\author{Shichao Sun\textsuperscript{\rm 1,5}, Ruifeng Yuan\textsuperscript{\rm 1}, Ziqiang Cao\textsuperscript{\rm 4}, Wenjie Li\textsuperscript{\rm 1}\thanks{\,\, Corresponding Authors. \\ \indent\indent   Work done while visiting GAIR Lab.}, Pengfei Liu\textsuperscript{\rm 2,3,5}\samethanks \\
   \textsuperscript{\rm 1} The Hong Kong Polytechnic University, \textsuperscript{\rm 2} Shanghai Jiao Tong University \\ 
   \textsuperscript{\rm 3} Shanghai Artificial Intelligence Laboratory,
   \textsuperscript{\rm 4} Soochow University   \\
    \textsuperscript{\rm 5} Generative AI Research Lab (GAIR)\\
  {\normalsize \texttt{cswjli@comp.polyu.edu.hk}, \texttt{pengfei@sjtu.edu.cn}}
}

\begin{document}
\maketitle
\begin{abstract}
Large language models (LLMs) have demonstrated the capacity to improve summary quality by mirroring a human-like iterative process of critique and refinement starting from the initial draft. Two strategies are designed to perform this iterative process: \textit{Prompt Chaining} and \textit{Stepwise Prompt}. Prompt chaining orchestrates the drafting, critiquing, and refining phases through a series of three discrete prompts, while Stepwise prompt integrates these phases within a single prompt. However, the relative effectiveness of the two methods has not been extensively studied. This paper is dedicated to examining and comparing these two methods in the context of text summarization to ascertain which method stands out as the most effective. Experimental results show that the prompt chaining method can produce a more favorable outcome. This might be because stepwise prompt might produce a simulated refinement process according to our various experiments. Since refinement is adaptable to diverse tasks, our conclusions have the potential to be extrapolated to other applications, thereby offering insights that may contribute to the broader development of LLMs.

\end{abstract}

\section{Introduction}
Large language models (LLMs) can enhance the summary via iterative refinement \citep{zhang-etal-2023-summit}. This is motivated by how humans refine their written text. The main idea contains three sequential steps: (1) \textbf{Drafting}: LLMs generate an initial summary;
(2) \textbf{Critiquing}: LLMs provide critical feedback and helpful suggestions for its output;
(3) \textbf{Refining}: LLMs use the feedback to refine the initial summary. 

More generally, this refinement can be applied to various text generation tasks to improve the outcomes \citep{madaan2023self, gou2023critic, selfee2023,akyurek-etal-2023-rl4f}. Moreover, the improved outcomes can also help train a more helpful and harmless model \citep{huang2022large,bai2022constitutional,OpenAI_GPT4_2023,scheurer2023training}. 
Implementing this refinement process can be approached in two distinct methods: \textit{Prompt Chaining}\footnote{Prompt chaining is introduced in \url{https://www.promptingguide.ai/techniques/prompt_chaining}} and \textit{Stepwise Prompt}.\footnote{Stepwise prompt is similar to specifying the steps required to complete a task at \url{https://platform.openai.com/docs/guides/prompt-engineering/tactic-specify-the-steps-required-to-complete-a-task}} Prompt chaining undertakes drafting, critiquing, and refining phases through a sequence of three discrete prompts. It means that LLMs will run three times. Although LLMs can concentrate on solving one particular problem without being overwhelmed by the complexity of the multiple tasks, it is trivial and troublesome for humans to provide three comprehensive prompts.
Conversely, stepwise prompt completes these three phases within a single generation. Stepwise prompt only needs a simple prompt to contain three sequential steps, but it is challenging for LLMs to generate a long and complex output. Currently, the effectiveness of these two methods remains underexplored in any text generation task. 

In this short paper, we compare prompt chaining and stepwise prompt to find the better method for refinement in text summarization. Specifically, we conduct experiments on the dataset \textbf{InstruSum} \citep{liu2023benchmarking} introduced to evaluate the capabilities of LLMs. It involves instruction controllable text summarization, which summarizes the article based on the specific requirement. We evaluate the quality of initial summaries, critiques, refined summaries to show the effect of prompt chaining and stepwise prompt. Experimental results indicate that the prompt chaining is better than stepwise prompt. Moreover, various experiments imply that \textbf{stepwise prompt might produce a simulated refinement process}, where LLMs intentionally produce errors only to subsequently correct them. Intuitively, this conclusion will work on other domains and further facilitate future research.

\begin{figure*}[ht]
    \centering
    \includegraphics[width=0.8\linewidth]{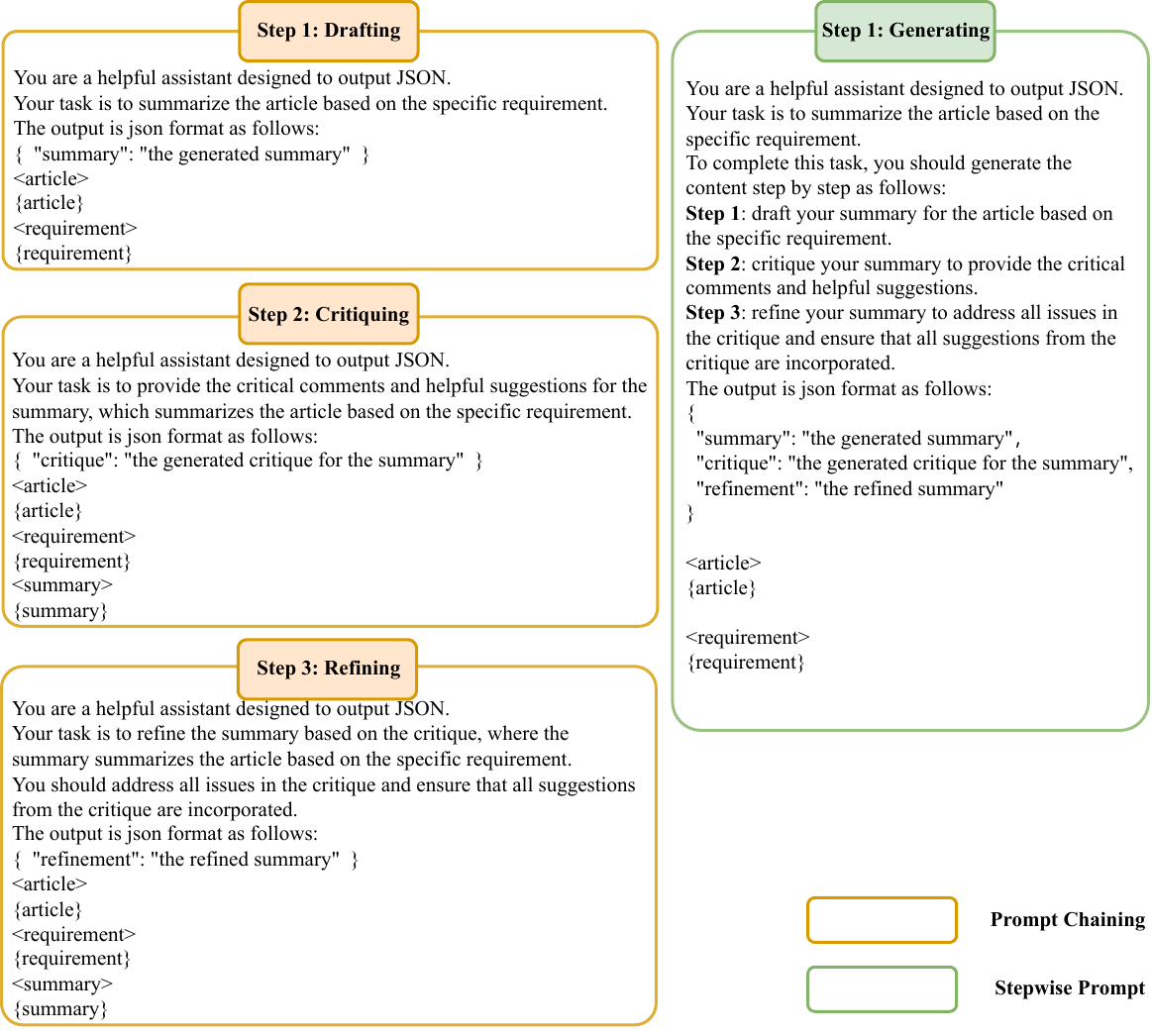}
    \caption{Prompt Chaining v.s. Stepwise Prompt.}
    \label{fig:prompts}
\end{figure*}

\section{Related Works}
Recent work has proved that refinement can significantly improve LLMs performance. 
Self-Refine \citep{madaan2023self} uses LLMs for drafting outcomes, providing feedback, and refining initial generation. In a series of 7 varied tasks, ranging from dialogue response to mathematical reasoning, outputs created using the Self-Refine method are favored over those produced through one-step generation with the same LLM, as judged by human evaluators and automated metrics.
Critic \citep{gou2023critic} proposes to leverage external tools to critique generated text and refine the initial generation via evaluation feedback. SelFee \citep{selfee2023} collects generations, feedback, and revised generations to finetune LLaMA models \citep{touvron2023llama}. \citet{akyurek-etal-2023-rl4f} propose to train a better critique model to help repair the model outputs. 
\citet{zhang-etal-2023-summit} introduce a refinement paradigm to enhance the faithfulness and controllability in text summarization. 
Moreover, the refined outcomes can also help train a more helpful and harmless model \citep{huang2022large,bai2022constitutional,OpenAI_GPT4_2023,scheurer2023training}. 

\section{Prompts}
Figure \ref{fig:prompts} illustrates the prompts of prompt chaining and stepwise prompt within the context of instruction controllable text summarization. Prompt chaining requires a human to segment the refinement process into three steps. Each step leverages the output from the preceding one. In contrast, stepwise prompt specifies the same three steps to be executed within a single operation. Therefore, they can generate the equivalent results, including: 
(1) \textbf{Draft Summary} is the initially generated summary. 
(2) \textbf{Critique} is the critical comment and the helpful suggestion.
(3) \textbf{Refined Summary} stems from refining the draft summary based on the critique. 
Correspondingly, these outcomes are obtained from each step in prompt chaining or the sequential items in the prompt chaining outcome.

\begin{table*}[ht]
\centering
\small
\setlength{\tabcolsep}{5pt}
\begin{tabular}{lcccccccccc}
\toprule
   \multirow{2}{*}{\textbf{Models}}       & \multicolumn{3}{c}{\textbf{Overall}} & \multicolumn{3}{c}{\textbf{Missing}} & \multicolumn{3}{c}{\textbf{Irrelevant}} \\ \cmidrule(lr){2-4} \cmidrule(lr){5-7} \cmidrule(lr){8-10}
          & \textbf{Win}          & \textbf{Tie}  & \textbf{Lose}       & \textbf{Win}          & \textbf{Tie}  & \textbf{Lose}   & \textbf{Win}          & \textbf{Tie}  & \textbf{Lose}  & \textbf{Length} \\ \midrule
\texttt{Mixtral-stepwise-draft} & 12         & 29        & 59       & 13  & 35        & 52   & 8 & 33 & 59 &  111.19  \\
\texttt{Mixtral-chaining-draft} & 18         & 27        & 55     & 19  & 41        & 40   & 11 & 46 & 43 & 119.63   \\  \midrule
\texttt{Mixtral-stepwise-refine} & 19        & 25        & 56       & 20  & 30       & 50  & 11 & 29 & 60 &  124.35  \\
\texttt{Mixtral-chaining-refine} & 27   & 21   & 52  & 31  & 29        & 40   & 14 & 48 & 38 &  127.3  \\ \midrule
\texttt{gpt-3.5-stepwise-draft} & 10         & 14        & 76       & 8  & 30        & 62   & 5 & 37 & 58 &  86.58  \\
\texttt{gpt-3.5-chaining-draft} & 12         & 22        & 66     & 13  & 28        & 59   & 7 & 37 & 56 & 94.76   \\ \midrule
\texttt{gpt-3.5-stepwise-refine} & 12        & 13        & 75       & 14  & 17       & 69  & 2 & 27 & 71 &  85.79  \\
\texttt{gpt-3.5-chaining-refine} & 21       & 17   & 62  & 14  & 24        & 62   & 11 & 38 & 51 &  97.24  \\ \midrule
\texttt{gpt-4-stepwise-draft}   & 34        & 40        & 26      & 27   & 53    & 20  & 16 & 60 & 24 & 125.73 \\ \midrule
\texttt{gpt-4-stepwise-refine}   & 53        &  29 & 18        &42  & 49    & 9 & 12& 57& 31& 145.85\\
\texttt{gpt-4-chaining-refine}   & \textbf{77}        &  14  & 9 & \textbf{57}  & 38        & 5 & \textbf{19} & 39 & 42 & 174.35\\
\bottomrule    
\end{tabular}
  \caption{Automatic benchmarking results. The summaries of different methods are compared against summaries generated by GPT-4 (\texttt{gpt-4-0125-preview}) one-step generation (i.e., \texttt{gpt-4-chaining-draft}) using the LLMCompare protocol \citep{liu2023benchmarking}. The average length of baseline summaries is 113.03.}
  \label{tab:benchmark}
\end{table*}

\section{Experiments and Results}

\subsection{Dataset}
We conduct experiments on the dataset \textbf{InstruSum} \citep{liu2023benchmarking}, which is produced to evaluate the capabilities of LLMs to summarize the article based on the specific requirement. InstruSum contains 100 article-requirement pairs in total. The articles contain around 1000-1200 words, stemming from the BBC news website.~\footnote{\url{
https://www.bbc.com/news}} Requirements for a summary are designed to reflect diverse information needs that readers may have at different stages of their reading journey. These requirements include (a) Informational requirement, which supplies pertinent details about the topic or subject being discussed within the articles; (b) Formatting requirement, which enhances the summary's structure, such as incorporating bullet lists, to improve its readability and facilitate quicker comprehension; (c) Meta requirement, which reflects a high-level overview of the article. 

\section{Models and Metrics}
Refinement can be powered by various LLMs. In this paper, we choose the newest versions of GPT-3.5 (\texttt{gpt-3.5-turbo-0125}) and GPT-4 (\texttt{gpt-4-0125-preview}) models from OpenAI \footnote{\url{https://platform.openai.com/docs/models}} to draft, critique, and refine the outcomes due to their strong instruction-following capabilities. We also explore the performance of a strong open-source LLM (Mixtral 8$\times$7B \citep{jiang2024mixtral}).

We use the LLMCompare as our evaluation protocol, which compares two candidate outputs and then selects the better one \citep{zheng2023judging,wang2023pandalm}. This is because LLMCompare coupled with GPT-4 is the best evaluation protocol, as mentioned in \citet{liu2023benchmarking}. The evaluation prompts are shown in Appendix \ref{sec:prompts}.

We evaluate the generated summaries from the three quality dimensions as introduced in \citet{liu2023benchmarking}: 
(1) \textbf{Overall Quality} measures the overall excellence of the summary following the summary requirements. 
(2) \textbf{Missing Information} assesses whether the summary omits any essential article details pertinent to the summary requirements.
(3) \textbf{Irrelevant Information} examines whether the summary contains extraneous information that falls outside the scope of the summary requirements. 

\subsection{Exp I: Summarization Benchmark}

\paragraph{Setup}
Consistent with the settings employed in previous research on automatic LLM benchmarking \citep{dubois2023alpacafarm,zheng2023judging}, we use GPT-4 (\texttt{gpt-4-0125-preview}) one-step outcomes as the baseline. We assess the performance of various methods through direct comparison with the baseline GPT-4 results. To mitigate the potential positional bias, the summary pairs are randomly shuffled before the evaluation. We perform the LLMCompare prompts via \texttt{gpt-4-0125-preview}. 

\paragraph{Results}
Table \ref{tab:benchmark} shows the automatic benchmarking results. More win times or less lose times mean a stronger performance. Generally, the draft summary is enhanced via refinement, regardless of the methods used. Notably, the performance of \texttt{gpt-3.5-stepwise-refine} (and \texttt{Mixtral-stepwise-refine}) is comparable to that of \texttt{gpt-3.5-chaining-draft} (and \texttt{Mixtral-stepwise-draft}). It indicates that stepwise prompt might lead to a simulated refinement process in which LLMs intentionally produce errors only to subsequently correct them.

\noindent\textbf{Q1: Which is the better method of prompt chaining and stepwise prompt?}

Prompt chaining achieves the highest win times (77 out of 100), considerably outshining stepwise prompt in producing higher-quality summaries. Moreover, prompt chaining coupled with a better backbone model can lead to better performance by comparing the outcomes of GPT 3.5 and GPT-4. 

\noindent\textbf{Q2: How does prompt chaining or stepwise prompt affect the initial outcome?}

Notably, summaries initially drafted using stepwise prompt frequently fall short in quality. This may be due to the anticipation that its outputs will subsequently undergo critique and refinement, potentially influencing the initial drafting process.

\begin{figure}[t]
\centering
\subfloat[\centering Overall]{{
\label{fig:overall}
\begin{tikzpicture}
\begin{axis}[
    xbar stacked,
    legend image code/.code={%
                    \draw[#1, draw=none] (0cm,-0.07cm) rectangle (0.2cm, 0.13cm);
    },  
    legend style={
        legend columns=3,
        at={(0.45,1)},
        anchor=south,
        draw=none,
        /tikz/every even column/.append style={column sep=0.3cm},
    },
    ytick=data,
    axis y line*=none,
    axis x line*=bottom,
    tick label style={font=\small},
    legend style={font=\small},
    label style={font=\small},
    xtick={0,25,50,75,100},
    width=0.4\textwidth,
    bar width=10pt,
    yticklabels={\texttt{average}, \texttt{gpt-4-0125},\texttt{gpt-4-1106}},
    xticklabels={0, 25\%, 50\%, 75\%, 100\%},
    xmin=0,
    xmax=100,
    y=5mm,
    enlarge y limits={abs=0.9},
    enlarge x limits={abs=0.2},
]
\addplot[rightred,fill=rightred] coordinates
{(39,0) (49,1) (29,2) };
\addplot[midgreen,fill=midgreen] coordinates
{(46,0) (31,1) (61,2) };
\addplot[leftgreen,fill=leftgreen] coordinates
{(15,0) (20,1) (10,2) };
\legend{Wins, Tie, Loses};
\coordinate (0 win) at (19,0);
\coordinate (0 tie) at (62,0);
\coordinate (0 lose) at (92,0);
\coordinate (1 win) at (24,5mm);
\coordinate (1 tie) at (65,5mm);
\coordinate (1 lose) at (90,5mm);
\coordinate (2 win) at (14,10mm);
\coordinate (2 tie) at (60,10mm);
\coordinate (2 lose) at (95,10mm);
\end{axis} 
\node at (0 win) {39};
\node at (0 tie) {46};
\node at (0 lose) {15};
\node at (1 win) {49};
\node at (1 tie) {31};
\node at (1 lose) {20};
\node at (2 win) {29};
\node at (2 tie) {61};
\node at (2 lose) {10};
\end{tikzpicture}
}}%

\subfloat[\centering Missing]{{
\label{fig:missing}
\begin{tikzpicture}
\begin{axis}[
    xbar stacked,
    ytick=data,
    axis y line*=none,
    axis x line*=bottom,
    tick label style={font=\small},
    label style={font=\small},
    xtick={0,25,50,75,100},
    width=0.4\textwidth,
    bar width=3mm,
    yticklabels={\texttt{average}, \texttt{gpt-4-0125},\texttt{gpt-4-1106}},
    xticklabels={0, 25\%, 50\%, 75\%, 100\%},
    xmin=0,
    xmax=100,
    y=5mm,
    enlarge y limits={abs=0.625},
    enlarge x limits={abs=0.2},
]
\addplot[rightred,fill=rightred] coordinates
{(25,0) (29,1) (21,2) };
\addplot[midgreen,fill=midgreen] coordinates
{(67.5,0) (61,1) (74,2) };
\addplot[leftgreen,fill=leftgreen] coordinates
{(7.5,0) (10,1) (5,2) };
\legend{};
\coordinate (0 win) at (12,0);
\coordinate (0 tie) at (62,0);
\coordinate (0 lose) at (96,0);
\coordinate (1 win) at (14,5mm);
\coordinate (1 tie) at (62,5mm);
\coordinate (1 lose) at (95,5mm);
\coordinate (2 win) at (10,10mm);
\coordinate (2 tie) at (64,10mm);
\coordinate (2 lose) at (98,10mm);
\end{axis} 
\node at (0 win) {25};
\node at (0 tie) {67.5};
\node at (0 lose) {7.5};
\node at (1 win) {29};
\node at (1 tie) {61};
\node at (1 lose) {10};
\node at (2 win) {21};
\node at (2 tie) {74};
\node at (2 lose) {5};
\end{tikzpicture}
}}%

\subfloat[\centering Irrelevant]{{
\label{fig:betterRhuman}
\begin{tikzpicture}
\begin{axis}[
    xbar stacked,
    ytick=data,
    axis y line*=none,
    axis x line*=bottom,
    tick label style={font=\small},
    label style={font=\small},
    xtick={0,25,50,75,100},
    width=0.4\textwidth,
    bar width=3mm,
    yticklabels={\texttt{average}, \texttt{gpt-4-0125},\texttt{gpt-4-1106}},
    xticklabels={0, 25\%, 50\%, 75\%, 100\%},
    xmin=0,
    xmax=100,
    y=5mm,
    enlarge y limits={abs=0.625},
    enlarge x limits={abs=0.2},
]
\addplot[rightred,fill=rightred] coordinates
{(18.5,0) (22,1) (15, 2) };
\addplot[midgreen,fill=midgreen] coordinates
{(62,0) (57,1) (67,2) };
\addplot[leftgreen,fill=leftgreen] coordinates
{(19.5,0) (21,1) (18,2) };
\legend{};
\coordinate (0 win) at (9,0);
\coordinate (0 tie) at (50,0);
\coordinate (0 lose) at (91,0);
\coordinate (1 win) at (11,5mm);
\coordinate (1 tie) at (48,5mm);
\coordinate (1 lose) at (89,5mm);
\coordinate (2 win) at (6,10mm);
\coordinate (2 tie) at (50,10mm);
\coordinate (2 lose) at (92,10mm);
\end{axis} 
\node at (0 win) {18.5};
\node at (0 tie) {62};
\node at (0 lose) {19.5};
\node at (1 win) {22};
\node at (1 tie) {57};
\node at (1 lose) {21};
\node at (2 win) {15};
\node at (2 tie) {67};
\node at (2 lose) {18};
\end{tikzpicture}
}}
\caption{Win rates of refined results from prompt chaining over stepwise prompt. The left-hand models are used to evaluate the refined outcome.}%
\label{fig:tasks} 
\end{figure}
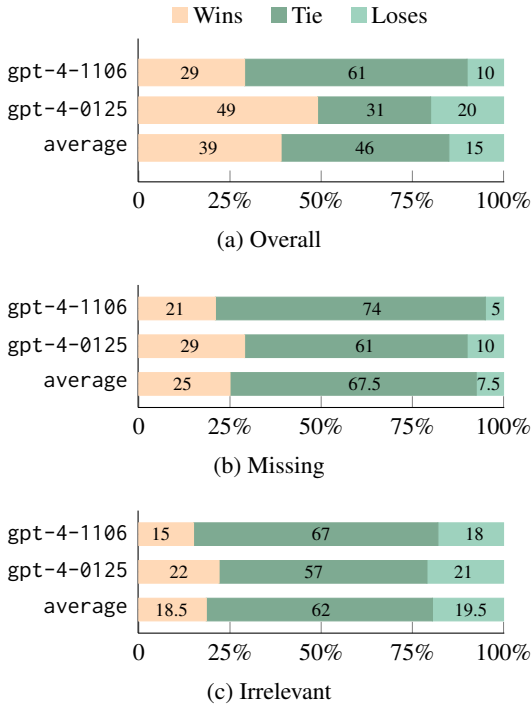

\subsection{Exp II: Robustness}
\paragraph{Setup}
Based on the understanding that different models used for LLMCompare evaluation can yield varied results as indicated by \citet{liu2023benchmarking}, we employ two iterations of the GPT-4 model, \texttt{gpt-4-1106-preview} and \texttt{gpt-4-0125-preview}, to validate the stability and robustness of prompt chaining's superiority over stepwise prompt. We do not use the GPT-3.5 models for powering LLMCompare evaluations due to their observed lower consistency with human evaluators. Lastly, \texttt{average} reports the mean value of the two scores.

\paragraph{Results}
Figure \ref{fig:tasks} shows the win rates between prompt chaining and stepwise prompt through refined results. 
The higher win rates of Overall suggest that prompt chaining more effectively adheres to the established summary requirements. 

\noindent\textbf{Q3: Does prompt chaining stably outperform stepwise prompt?}

We observe that prompt chaining beats stepwise prompt in both Overall and Missing evaluation across different evaluation models. Meanwhile, prompt chaining exhibits comparable performance to stepwise prompt in Irrelevant. It can confirm the reliability of our conclusion that prompt chaining stably outperforms stepwise prompt.

\begin{table*}[!htbp]
\centering
\small
\setlength{\tabcolsep}{4mm}
\begin{tabular}{lcccccccccc}
\toprule
   \multirow{2}{*}{\textbf{Models}}       & \multicolumn{3}{c}{\textbf{Overall}} & \multicolumn{3}{c}{\textbf{Missing}} & \multicolumn{3}{c}{\textbf{Irrelevant}} \\ \cmidrule(lr){2-4} \cmidrule(lr){5-7} \cmidrule(lr){8-10}
          & \textbf{Win}          & \textbf{Tie}  & \textbf{Lose}       & \textbf{Win}          & \textbf{Tie}  & \textbf{Lose}   & \textbf{Win}          & \textbf{Tie}  & \textbf{Lose}   \\ \midrule
\texttt{GPT 3.5} & 16         & 5        & 9       & 15  & 7        & 8   & 6 & 20 & 4   \\
\texttt{GPT 4} & 14         & 8        & 8     & 13 & 10        & 7   & 9 & 14 & 7    \\ 
\texttt{Mixtral} & 11        & 16        & 3       & 7  & 22       & 1  & 6 & 19 & 5  \\
\bottomrule    
\end{tabular}
  \caption{Human evaluation results. }
  \label{tab:human_eval_s}
\end{table*}

\section{Exp III: Human Evaluation}
\paragraph{Setup}
We engaged two postgraduate students to conduct human evaluation, wherein they compared the refined outcomes of Prompt Chaining against those of the Stepwise Prompt. If Prompt Chaining outperforms Stepwise Prompt, it is notated as a ``Win''. For this human evaluation, we randomly selected 30\% data from InstruSum dataset. Similar to the automated evaluation, we also use ``overall'', ``missing'', ``irrelevant'' as the evaluation metrics. 

\paragraph{Results}

Table \ref{tab:human_eval_s} presents the quality of critique. A higher score means a better performance. The ``win'' times significantly exceed the ``los'' times. It indicates that prompt chaining outperforms stepwise prompt. This conclusion is consistent with GPT-4 automated evaluation. Additionally, we observe that there are fewer ``lose'' times when we apply the more advanced model, GPT-4. It may imply that Prompt Chaining significantly outperforms Stepwise Prompt when using advanced models.

\section{Exp IV: Critique Evaluation}
\paragraph{Setup}
We use \textsc{MetaCritique} \citep{sun2024critique} powered by \texttt{gpt-4-0613} to evaluate the quality of critiques, which are the intermediate outputs of prompt chaining and stepwise prompt. \textsc{MetaCritique} involves three metrics: (1) \textbf{Precision} gauges the factuality of the critique; (2) \textbf{Recall} measures the comprehensiveness of the critique; (3) \textbf{F1 Score} harmonizes the precision score and recall score. We do not assess GPT-4 critiques, as \textsc{MetaCritique} uses GPT-4 outcomes as references.

\paragraph{Results}

Table \ref{tab:critique_eval} presents the quality of critique. A higher score means a better performance. 

\noindent\textbf{Q4: How does prompt chaining or stepwise prompt affect the critique generation?}

Stepwise prompt can generate high-quality critiques that are both more factual and comprehensive. 
However, in terms of F1 score, prompt chaining achieves only half of that of stepwise prompt, despite the superior performance in refined summaries. These results imply that stepwise prompt produces a simulated refinement process.

\begin{table}[!htbp]
\centering
  \small
   \setlength{\tabcolsep}{5pt}
\begin{tabular}{lccc}
\toprule
    \multirow{2}{*}{\textbf{Models}}    & \multicolumn{3}{c}{\textbf{MetaCritique}} \\ \cmidrule(lr){2-4}
            &   \textbf{Precision}    &     \textbf{Recall}  & \textbf{F{\small 1} Score }        \\ \midrule
\texttt{gpt-3.5-stepwise}  &   78.91     &   43.29   &  52.48  \\
\texttt{gpt-3.5-chaining}   &   40.21   &   25.62  & 24.79   \\  
\bottomrule    
\end{tabular}
  \caption{\textsc{MetaCritique} scores. }
  \label{tab:critique_eval}
\end{table}

\section{Conclusion}
LLMs can enhance summaries by emulating the human-like process of critique and refinement of their initial drafts. 
This paper explores two distinct strategies for implementing this process: \textit{Prompt Chaining} and \textit{Stepwise Prompt}.
We conduct rigorous experiments in the context of text summarization. Our findings indicate that prompt chaining garners a superior performance. Besides, the results imply that stepwise prompt might produce a simulated refinement process. Given that such refinement can be adapted to various tasks, our insights could extend beyond text summarization, potentially advancing the progress of LLMs. 

\section*{Acknowledgements}
We thank the anonymous reviewers for their valuable feedback and helpful suggestions. This project is supported by Research Grants Council of Hong Kong (PolyU/15203617 and PolyU/5210919), National Natural Science Foundation of China (62076212 and 62106165), Qingyuan Research Project and Shanghai Artificial Intelligence Laboratory.

\section*{Limitations}
Refinement can be applied to various natural language processing (NLP) tasks. However, this paper only compares prompt chaining and stepwise prompt in the scope of text summarization. Future research is warranted to validate the effectiveness of these strategies on an expansive range of NLP tasks, thereby enhancing the generalizability of our findings and their potential utility across the field.

\section*{Ethical Considerations}

Our experimental data stems from InstruSum, which is well-established and publicly available. Dataset construction and annotation are consistent with the intellectual property and privacy rights of the original authors. This work complies with the ACL Ethics Policy\footnote{\url{https://www.aclweb.org/portal/content/acl-code-ethics}}. 

\bibliography{anthology,custom}

\appendix
\section{Prompts}
\label{sec:prompts}
We elaborate on the prompts for GPT-4 evaluation in Table \ref{tab:compare_overall}, \ref{tab:compare_missing} and \ref{tab:compare_irrelevant} for LLMCompare Overall, LLMCompare Missing, and LLMComapre Irrelevant.

\begin{table*}[ht]
    \scriptsize
    \centering
\begin{tabular}{@{}p{\textwidth}@{}}
\toprule
--------------SYSTEM MESSAGE-------------

~

You are a helpful assistant designed to output JSON.

In this task, you will be provided with a news article, a specific summary requirement, and two summaries.
The summaries are crafted to meet a specific summary requirement. Note that there may be identical summaries.

Your task is to compare the overall quality of these two summaries concerning the summary requirement and pick the one that is better (there can be a tie).

First you will give an explanation of your decision then you will provide your decision in the format of 1 or 2 or tie.

Please refer to the example below for the format of your response.

Example Response:

\{

  \quad "explanation": "Your explanation here",
  
  \quad "decision": "1 or 2 or tie",

\}

~

--------------USER MESSAGE-------------

~

<article>

\{article\}

<requirement>

\{requirement\}

<summary 1>

\{summary 1\}

<summary 2>

\{summary 2\}

\\
\bottomrule
\end{tabular}
    \caption{Prompt for LLMCompare Overall.}
    \label{tab:compare_overall}
\end{table*}

\begin{table*}[ht]
    \scriptsize
    \centering
\begin{tabular}{@{}p{\textwidth}@{}}
\toprule
--------------SYSTEM MESSAGE-------------

~

You are a helpful assistant designed to output JSON.

In this task, you will be provided with a news article, a specific summary requirement, and two summaries.
The summaries are crafted to meet a specific summary requirement. Note that there may be identical summaries.

Your task is to compare the quality of these two summaries concerning whether they omit any crucial information from the article with respect to the summary requirement and pick the one that is better (there can be a tie). Crucial information refers to key details or facts that are essential to understanding the article and meeting the summary requirement.

First you will give an explanation of your decision then you will provide your decision in the format of 1 or 2 or tie.

Please refer to the example below for the format of your response.

Example Response:

\{

 \quad "explanation": "Your explanation here",
 
 \quad "decision": "1 or 2 or tie",
 
\}

~

--------------USER MESSAGE-------------

~

<article>

\{article\}

<requirement>

\{requirement\}

<summary 1>

\{summary 1\}

<summary 2>

\{summary 2\}

\\
\bottomrule
\end{tabular}
    \caption{Prompt for LLMCompare Missing.}
    \label{tab:compare_missing}
\end{table*}

\begin{table*}[ht]
    \scriptsize
    \centering
\begin{tabular}{@{}p{\textwidth}@{}}
\toprule
--------------SYSTEM MESSAGE-------------

~

You are a helpful assistant designed to output JSON.

In this task, you will be provided with a news article, a specific summary requirement, and two summaries.
The summaries are crafted to meet a specific summary requirement. Note that there may be identical summaries.

Your task is to compare the quality of these two summaries concerning whether they include any information that is not relevant to the summary requirement and pick the one that is better (there can be a tie).
First you will give an explanation of your decision then you will provide your decision in the format of 1 or 2 or tie.

Please refer to the example below for the format of your response.

Example Response:

\{

  \quad "explanation": "Your explanation here",
  
  \quad "decision": "1 or 2 or tie",

\}

~

--------------USER MESSAGE-------------

~

<article>

\{article\}

<requirement>

\{requirement\}

<summary 1>

\{summary 1\}

<summary 2>

\{summary 2\}

\\
\bottomrule
\end{tabular}
    \caption{Prompt for LLMCompare Irrelevant.}
    \label{tab:compare_irrelevant}
\end{table*}

\end{document}